\documentclass[journal]{IEEEtran}
\usepackage{graphicx}
\usepackage{float}
\usepackage{subcaption}
\usepackage{amsmath}
\usepackage{bm}
\usepackage{upgreek}
\usepackage[numbers,sort&compress]{natbib}
\usepackage{color}

\ifCLASSINFOpdf
\else
\fi
\begin{document}

\title{A Fuzzy Logic-based Cascade Control without Actuator Saturation for the Unmanned Underwater Vehicle Trajectory Tracking}
%
%
%

\author{Danjie~Zhu,~Simon~X.~Yang~and~Mohammad~Biglarbegian
\thanks{This work was supported by the Natural Sciences and Engineering Research Council (NSERC) of Canada. \textit{(Corresponding author: Simon X. Yang.)} The authors are with Advanced Robotics and Intelligent System (ARIS) Laboratory  and Advanced \& Intelligent Control for Vehicles (AICV) Laboratory, School of Engineering, University of Guelph, Guelph, ON. N1G2W1, Canada (e-mail: \{danjie; syang; mbiglarb\}@uoguelph.ca).}
}

\maketitle

\begin{abstract}
An intelligent control strategy is proposed to eliminate the actuator saturation problem that exists in the trajectory tracking process of unmanned underwater vehicles (UUV). The control strategy consists of two parts: for the kinematic modeling part, a fuzzy logic-refined backstepping control is developed to achieve control velocities within acceptable ranges and errors of small fluctuations; on the basis of the velocities deducted by the improved kinematic control, the sliding mode control (SMC) is introduced in the dynamic modeling to obtain corresponding torques and forces that should be applied to the vehicle body. With the control velocities computed by the kinematic model and applied forces derived by the dynamic model, the robustness and accuracy of the UUV trajectory without actuator saturation can be achieved. 
\end{abstract}

%
\IEEEpeerreviewmaketitle

\section{Introduction}
\IEEEPARstart
{T}{o} take advantages of the abundant resources embedded in the ocean area, such as mineral resources, biological resources and space resources, technologies that relate to the underwater exploration have been studied for decants \cite{r4}. Due to the complex environmental factors of the deep-water space, such as the high pressure, invisibility or the unpredictable obstacles, unmanned underwater vehicles (UUV) are applied in most undermarine 
operation cases to guarantee the safety and efficiency \cite{r5,TvT2}. Therefore, achieving the robustness and accuracy of controlling the UUV to track the desired trajectory in the desired time, is of great importance for completing the underwater operation \cite{r7,TVT1,Wang2020JIRS}. 

However, as the UUV cannot provide infinite driving inputs such as torques/forces due to its underwater workspace and limited electric power, the actuator saturation has to be considered during the trajectory tracking process of the vehicle\cite{r9, r10, ry2}, with the torques/forces constraints applied. The actuator saturation is induced by the speed-jump problem, which usually occurs in some conventional control methods for trajectory tracking, like the backstepping control \cite{r8,Sanchez2021JIRS}. The speed jumps negatively affects the robustness of the UUV trajectory tracking, by introducing excessive fluctuations of velocities at initial states or other large error states during the kinematic controlling procedure. In the backstepping method, control functions for each subsystem are designed based on the Lyapunov techniques, and generated to form the complete control law \cite{r28}. Therefore speeds of large fluctuations are derived by the large errors accumulated from the generation of the subsystems, where speed-jump issues are induced when the deviation occurs. 

Many methods have been used for alleviating the speed-jump problem in the trajectory tracking for vehicles, where the model predictive control (MPC) is one of the most commonly applied methods. The MPC resolves the online optimization problem at each time step and derives in-time predictions with minimum errors \cite{mpc1,TvT3}. However, MPC usually consumes long time in computation due to its recursive algorithm with increasing complexity \cite{mpc2}. In this study, the fuzzy logic are introduced to provide low complexity by translating the goals in a transparent way \cite{fuz1}. The fuzzy logic system is used as the function approximator to address the uncertainties, and gives more flexiable criterion for obtaining the optimized predictions within its conceptual framework \cite{Jing,Fang}. It can also makes limitation on the output data, and smoothen the kinematic error curves derived by the conventional backstepping method through its decision function. Compared to the MPC, the fuzzy logic controller constructs a model that imitates the human decisionmaking with inputs of continuous values between 0 and 1, which largely simplifies the computing process \cite{Lee1,Lee2}.

Practically, as UUV driving commands are directly given by the dynamic inputs, the component that extents the kinematic to dynamic tracking is cascaded as a part of the controller designed in this study \cite{r11,r12}. Tiny deviation caused by the speed-jump problem leads to the inevitable errors in the dynamics of the tracking process, where the UUV may produce excessive torques/forces at the jump points. The controller designed for the dynamic model is to compute the corresponding torques/forces that directly applied to the vehicle to eliminate the errors created during the tracking procedure, which offers an accurate operating instruction to the diving vehicle \cite{r12,r13,ry1}. In this paper, the sliding mode control (SMC) is chosen to construct the complete intelligent controller \cite{Agu2022, Rubio2022, Agu2021,Soria,Soriano2020,Silva2021}. As one of the most basic adaptive controlling strategies, the SMC is widely used due to its simple and robust mechanism \cite{rcn1,rcn2}. A surface mode is supposed to follow the desired tracking and keep the control outputs remain on the surface. Once the trajectory under the control is out of the perfect surface, the SMC will push the trajectory slide back to the surface with addition or subtraction on the original controlling equation \cite{r17,r18}. Therefore the SMC restricts the fluctuation of control outputs in an acceptable range through a simple operation, which is highly applicable in the trajectory tracking problems\cite{r19,IJC2}. 

Zhang and his colleagues have tried to resolve the speed-jump problem in the trajectory tracking, but higher complexities are introduced \cite{Zhang}. Some researchers have achieved successful tracking based on the fuzzy logic-refined backstepping method yet their application is based on the underactuated surface vehicle (USV), with fewer states involved compared to the UUV \cite{Ning,Wang}. Some researchers have applied synergetic learning in their controller designed for vehicles and better performance is obtained, but they do not consider practical constraints of the vehicle \cite{tdcs_synergetic}. Li has developed the fuzzy logic-based controller that provides satisfactory tracking results even with time-varying delays or input saturation, but the effectiveness of the algorithm on specific models such as the UUV has not been discussed \cite{Yongming,yongming2} Wang and his colleagues developed a fuzzy logic-based backstepping method yet it has not been experimented under specific application scenarios, with dynamic constraints applied \cite{r22}. 

Motivated by the requirement of resolving the actuator saturation (thrusters' dynamic constraints) through the elimination of the speed jumps that exist in the conventional trajectory control of the UUV, this paper focuses on the speed-jump as well as the actuator saturation problems in the practical UUV system. Due to the uncertainty of the underwater environment, high adaptiveness and low complexity are needed to achieve a robust trajectory tracking controller that is easy to realize. Therefore, the fuzzy logic, the backstepping method and the SMC are combined to construct a cascade intelligent control. The first two components form the kinematic velocity controller, where the fuzzy logic system helps to resolve the speed-jump problem of the backstepping method when controlling the kinematic model. The SMC is constructed as the dynamic torque controller, extending the application of the whole design for UUV trajectory tracking in actual cases, meanwhile the physical constraints can be introduced in this part. Based on the shunting characteristics of this control strategy, the outputs are bounded in a finite interval within the vehicle's physical constraints and results of small fluctuations are performed even when abrupt inputs are given. The contribution of the cascade control method is supposed to resolve the actuator saturation problem and provides satisfactory tracking results in practical cases of UUV navigation through a simple computation. Moreover, the problem of navigation under uncertainties in stochastic environments is considered due to their impacts on the vehicle motion \cite{JIRS1,JIRS2}.

The rest of the paper is organized as follows. First, the basic models of the UUV system are introduced, kinematic model and dynamic model are defined with their corresponding formulas. The specific modeling process shows how the trajectory tracking control works in the complex system. Next, the fuzzy logic-refined backstepping control and the sliding mode control designed for the UUV trajectory tracking problem are illustrated, where the mechanism and operating process are explained in details. The final part presents the direct results output by the simulation, and further analysis is performed to demonstrate the effectiveness of the refined trajectory tracking controller with dynamic constraints applied and environmental noise involved.

 
\section{ROBOT MODELS AND PROBLEM STATEMENT}
In this section, a typical type of UUV named "Falcon" is studied. Its robot models and trajectory tracking problem descriptions are given in the form of specific equations . 
\subsection{Robot models of the "Falcon UUV"}
In this subsection, the kinematic and dynamic models of the “Falcon” UUV are given, both of which are involved in the trajectory tracking control of a UUV. Parameters of the “Falcon” UUV are introduced to clearly address the trajectory tracking problem and its corresponding solution studied in this paper.
\subsubsection{Kinematic Model}
The systematic analysis of UUV is established on two basic 3D reference frames, the world-fixed frame (W), originating from a point on the surface of the earth; and the body-fixed frame, originating from the UUV body. Directions of axes of the two reference frames are given in Fig. 1. Among the six freedoms of the UUV, surge, sway, heave, roll, pitch and yaw, roll and pitch can be eliminated when establishing the trajectory model to keep a controllable operation of the UUV during the diving process. 
Specially for the UUV type “Falcon” applied in this article, the design of the vehicle does not allow the roll and pitch movements while only surge, sway, heave and yaw movements can be achieved (see bold DOFs in Fig. 1) \cite{r24}. Therefore, for the kinematic equation of “Falcon” UUV, the velocity vector $\mathbf{v}$ can be transformed into the time derivative of trajectory vector
 $\mathbf{\dot{p}}$ as
$$
   \mathbf{\dot{p}}= 
 \begin{bmatrix}
   \dot{x} \\
   \dot{y} \\
   \dot{z} \\
   \dot{\psi}
  \end{bmatrix}
=\mathbf{J}(\mathbf{p})\mathbf{v}=
\begin{bmatrix}
   \cos{\psi}& -\sin{\psi} & 0 & 0 \\
   \sin{\psi}& \cos{\psi} & 0 & 0 \\
   0 & 0 & 1 & 0 \\
   0 & 0 & 0 & 1
  \end{bmatrix}\mathbf{v}\\$$
  \begin{equation}
  =\begin{bmatrix}
   \cos{\psi}& -\sin{\psi} & 0 & 0 \\
   \sin{\psi}& \cos{\psi} & 0 & 0 \\
   0 & 0 & 1 & 0 \\
   0 & 0 & 0 & 1
  \end{bmatrix}
  \begin{bmatrix}
   u \\
   v \\
   w \\
   r
  \end{bmatrix}, 
\end{equation} 
where $\mathbf{J}$ is a transformation matrix derived from the physical structure of the UUV body, $[u~v~w~r]^T$ represents the velocities at the chosen four axes of the UUV (see Fig. 1) \cite{mthesis1}.
\begin{figure}[htbp]
\begin{center}
    \includegraphics[scale=0.3]{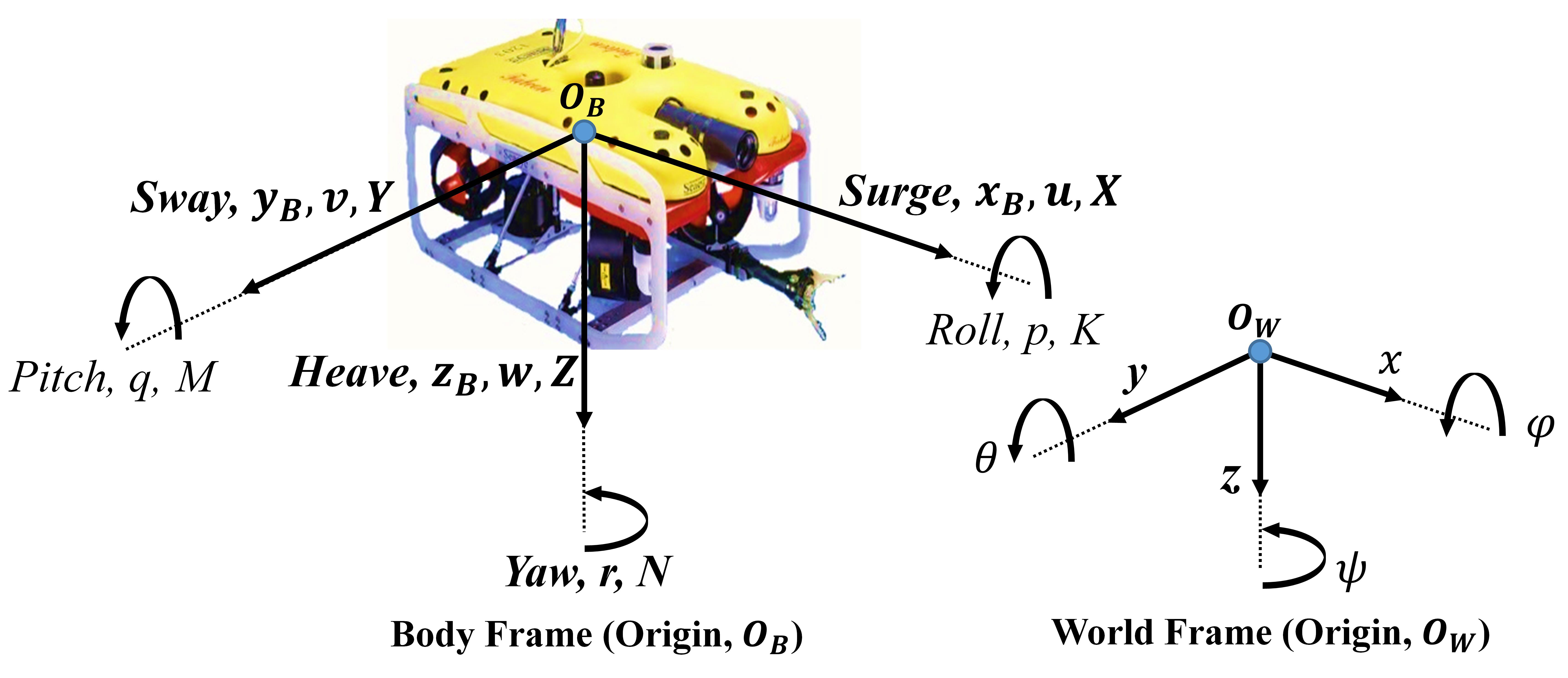}                         
    \caption*{Fig. 1. Reference frames and six degrees of freedom (x, y, z, $\theta$, $\phi$ and $\psi$) of a UUV}	
\end{center} 
\end{figure}
\subsubsection{Dynamic Model}
In an actual UUV system, several complex and nonlinear forces such as hydrodynamic drag, damping, lift forces, Coriolis and centripetal forces, gravity and buoyancy forces, thruster forces, and environmental disturbances are acting on the vehicle. Considering the origins and effect of the forces, a general dynamic model can be written as
\begin{equation}
\mathbf{M}\mathbf{\dot{v}}+\mathbf{C}(\mathbf{v})\mathbf{v}+\mathbf{D}(\mathbf{v})\mathbf{v}+\mathbf{g}(\mathbf{p})=\boldsymbol{\uptau}\,,
\end{equation}
where M is the inertia matrix of the summation of rigid body  and added mass; $\mathbf{C}(\mathbf{v})$ is the Coriolis and centripetal matrix of the summation of rigid body and added mass; $\mathbf{D}(\mathbf{v})$ is the quadratic and linear drag matrix; $\mathbf{g}(\mathbf{p})$ is the matrix of gravity and buoyancy; and $\boldsymbol{\uptau}$ is the torque vector of the thruster inputs. 
As is mentioned in the previous section, in this study only four states are considered for the specific model Falcon UUV. All the matrixes listed in the dynamic model will be the matrix $\in$ $R^{4 \times 4}$, and the vector $\in$ $R^{4 \times 1}$. The torque vector of the thruster input is represented by
\begin{equation}
    \boldsymbol{\uptau}=
    \begin{bmatrix}
       \tau_x & \tau_y &
       \tau_z &
       \tau_\psi
    \end{bmatrix},
\end{equation}
where $x$, $y$ and $z$ representing the linear displacements of the UUV at surge, sway and heave directions, while $\psi$ representing the angular displacement of the UUV at yaw direction (see Fig. 1). 
\subsubsection{Torque-force Transition and Normalization}
The four torques applied to the four different states of the Falcon UUV are derived from the five thrusters that distributed on the UUV body. In the simplified structure of Falcon UUV presented in Fig. 2, four horizontal thrusters are set on the edges of the vehicle to achieve displacements at surge, sway and yaw directions, and one vertical thruster is set in the center above the vehicle to realize the displacement at heave direction.  
\begin{figure}[ht]
\begin{center}
    
        \includegraphics[scale=0.6]{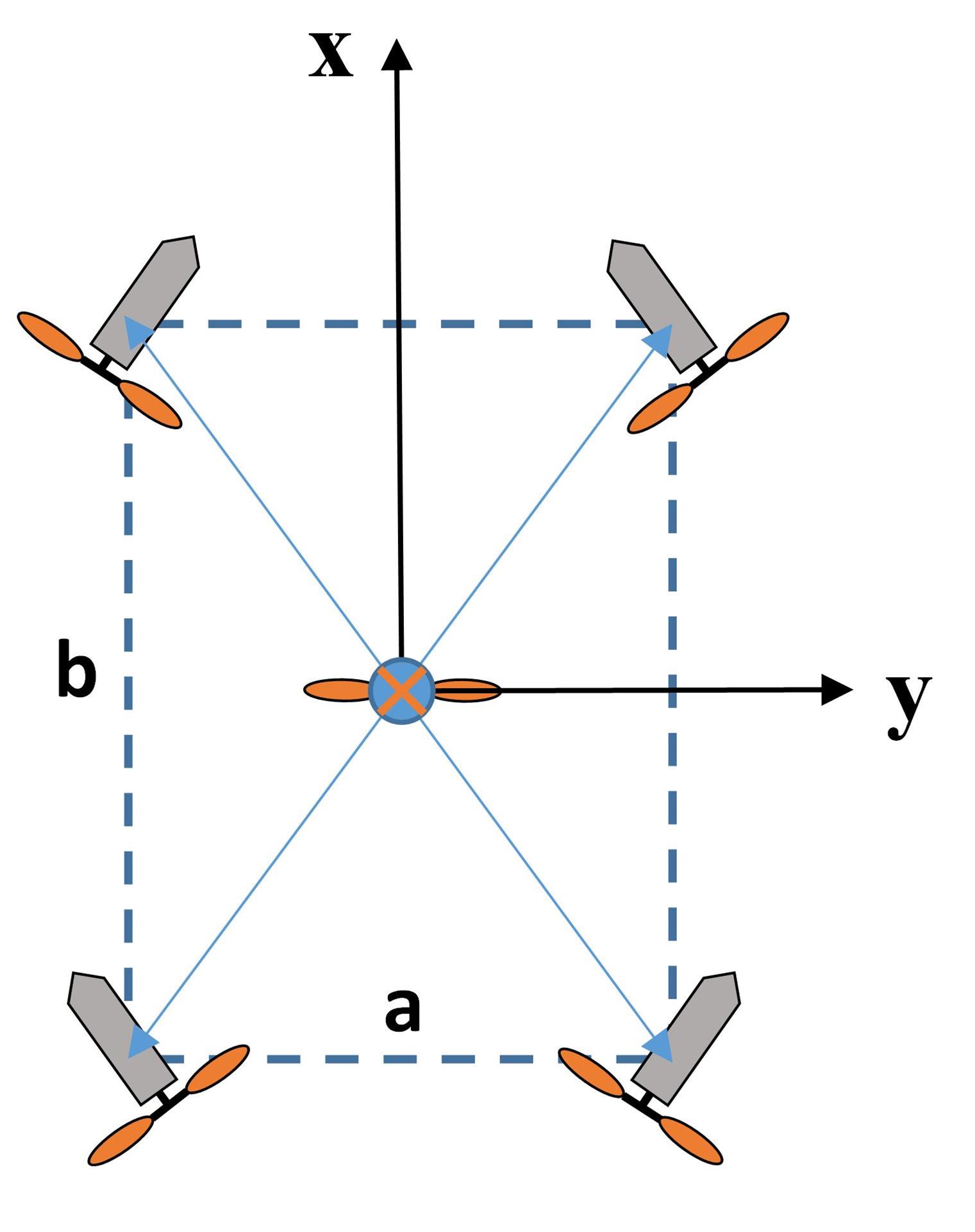}                         
        \caption*{Fig. 2. Distribution of the five thrusters on the body of the “Falcon” UUV}	
    
\end{center} 
\end{figure}
As the forces applied to the five thrusters shown in Fig. 2, the torques at the four different states can be calculated as the following, supposing $A=\frac{a}{2}\cos{\alpha}+\frac{b}{2}\sin{\alpha}$,
\begin{equation}
    \begin{bmatrix}
       \tau_x\\
       \tau_y\\
       \tau_z\\
       \tau_\psi
    \end{bmatrix}=
    \begin{bmatrix}
       T_1\cos{\alpha}+T_2\cos{\alpha}+T_3\cos{\alpha}+T_4\cos{\alpha}\\
       T_1\sin{\alpha}-T_2\sin{\alpha}+T_3\sin{\alpha}-T_4\sin{\alpha}\\
       T_5\\
       AT_1-AT_2+AT_3-AT_4
    \end{bmatrix},
\end{equation}
where $T_1$, $T_2$, $T_3$, $T_4$ are the forces applied to the four thrusters arranged on the edges, $T_5$ is the force applied to the center thruster that controls the vertical depth. The transition between torque vector and force vector can be written as
\begin{equation}
    \begin{bmatrix}
       \tau_x\\
       \tau_y\\
       \tau_z\\
       \tau_\psi
    \end{bmatrix}=
    \begin{bmatrix}
       \cos{\alpha} & \cos{\alpha} & \cos{\alpha} & \cos{\alpha} & 0\\
       \sin{\alpha} & -\sin{\alpha} & \sin{\alpha} & -\sin{\alpha} & 0\\
       0 & 0 & 0 & 0 & 1\\
       A & -A & -A& A & 0
    \end{bmatrix}
    \begin{bmatrix}
       T_1\\
       T_2\\
       T_3\\
       T_4\\
       T_5
    \end{bmatrix}.
\end{equation}

If the five thrusters of the same type are supposed to have a maximum force $T_m$, the maximum of the torque vector $\boldsymbol{\uptau_m}$ can be further deducted as
\begin{equation}
    \boldsymbol{\uptau_m}=
    \begin{bmatrix}
       \tau_{xm}\\
       \tau_{ym}\\
       \tau_{zm}\\
       \tau_{\psi m}
    \end{bmatrix}=
    \begin{bmatrix}
       4T_m\cos{\alpha}\\
       4T_m\sin{\alpha}\\
       T_m\\
       4T_mA
    \end{bmatrix}.
\end{equation}

Substitute into Eq. (5), and divide both sides by the maximum torques to restrict the output in a certain range of -1 to 1, the vector is transformed into
$$ \begin{bmatrix}
       \tau_x/\tau_{xm}\\
       \tau_y/\tau_{ym}\\
       \tau_z/\tau_{zm}\\
       \tau_{\psi}/\tau_{\psi m}
    \end{bmatrix}\\
    =
    \begin{bmatrix}
       1/4 & 1/4 & 1/4 & 1/4 & 0\\
       1/4 & -1/4 &1/4 & -1/4 & 0\\
       0& 0&0 &0&1\\
       1/4 & -1/4& -1/4 & 1/4 & 0
    \end{bmatrix}
    \begin{bmatrix}
       T_1/T_m\\
       T_2/T_m\\
       T_3/T_m\\
       T_4/T_m\\
       T_5/T_m
    \end{bmatrix}\\
$$  \begin{equation} 
    =
    \overline{\mathbf{B}} \begin{bmatrix}
       T_1/T_m\\
       T_2/T_m\\
       T_3/T_m\\
       T_4/T_m\\
       T_5/T_m
    \end{bmatrix}.
\end{equation}

After further simplification, set $\overline{\boldsymbol{\uptau}}=\boldsymbol{\uptau}/\boldsymbol{\uptau_m}$, $\mathbf{\overline{T}}=\mathbf{T}/\mathbf{T_m}$, then the above equation has the compact form as
\begin{gather}
\overline{\boldsymbol{\uptau}}=\overline{\mathbf{B}}\,\mathbf{\overline{T}}\notag
\\
\,\mathbf{\overline{T}}=\overline{\mathbf{B}}^{-1}\, \overline{\boldsymbol{\uptau}}\,,
\end{gather}
where $\overline{\mathbf{B}}^{-1}$ is the generalized inverse matrix of $\overline{\mathbf{B}}$.

Therefore, the transition between forces applied to the thrusters and the torques is achieved and normalized. For all the torques and forces in Eq. (8), they are all ranged from -1 to 1 to perform a direct and simplified showcase during the tracking control process.

\subsection{Problem Statement}
In this subsection, the trajectory tracking problem of the "Falcon" UUV is described and physicial constraints (actuator saturation) of the UUV is introduced.
\subsubsection{Trajectory Tracking Problem of the UUV}
  To realize the trajectory tracking of the UUV, the vehicle must follow the desired path along the corresponding time period. In other words, the errors between the desired and actual trajectories have to be minimized at the different degrees of freedom \cite{r26}. In the kinematic model of UUV, ideal trajectory tracking can be realized by
\begin{equation}
\mathbf{e}(t)=\mathbf{p_d}(t)-\mathbf{p}(t)\,\rightarrow\,0\,,
\end{equation}
where $\mathbf{p_d}(t)$ is a vector representing the desired trajectory of the UUV, $\mathbf{p}(t)$ is a vector representing the actual trajectory, and $\mathbf{e}(t)$ is a vector of errors between the desired and actual trajectories at the four axes, shown in the following form as
\begin{equation}
\mathbf{e}(t)=
\begin{bmatrix}
   e_x\\
   e_y\\
   e_z\\
   e_\psi
\end{bmatrix}=
\begin{bmatrix}
   x_d-x\\
   y_d-y\\
   z_d-z\\
   \psi_d-\psi
\end{bmatrix}.
\end{equation}

Therefore, when designing the trajectory tracking controller, the error variables should be as small as possible to obtain good tracking results.
\subsubsection{Constraints on Trajectory Tracking Control of the UUV}
In the actual application, the trajectory tracking effect is always restricted by physical constraints of the vehicle. According to the practical construction and application of the unmanned underwater vehicle, the UUV cannot provide infinite driving inputs such as torques/forces to complete the navigation, thus resulting into the problem of actuator saturation. The removal of the dynamic constraints might be realized under the condition where the power can be provided unboundedly and the model parameters are fully detected. This condition is temporarily hard to realize for the UUV as the underwater environment is highly unpredictable, as well as the actual movement of the vehicle. Therefore driving restrictions are always applied on the UUV to achieve a reliable and controllable navigation process. Based on the UUV structure, bounds of UUV's output velocities are restricted by forces produced by the vehicle at different axes (see Fig. 1), which are provided by its thrusters physically (see Fig. 2). Hence the maximum forces of the vehicle thrusters, derived by the maximum torques that can be offered by the vehicle body, are the essential constraints of the UUV tracking problem.

To assess the influence of the constraints, maximum torques $\boldsymbol{\uptau_m}$ are introduced in the simulation part of this paper. By the definition given in Eq. (8), normalized torques $\boldsymbol{\overline{\uptau}}$ and normalized forces $\mathbf{\overline{T}}$ are supposed to have the limitation of -1 to 1. The two variables are used to quantify the effect of the constraints during the trajectory tracking process in the simulation part. 

\section{Fuzzy-Refined Backstepping with Sliding mode control Trajectory Tracking (FBSTT) Design}
The basic control architecture of the system is illustrated in Fig. 3. The design of the control strategy consists of two parts: (1) an outer loop of auxiliary kinematic control based on the positions and orientation state errors of the UUV; (2) an inner loop of sliding mode control based on the velocity state vector. The kinematic/dynamic cascade control system integrates backstepping technology and sliding mode control together with a fuzzy logic specially designed in this study. The environmental disturbance is also considered in the second part of the simulation, which is addressed by introducing a random error input and filtered by the sensors, at the status of forming the trajectory. The details of the control design will be presented in the later section.
\begin{figure*}[ht]
\begin{center}
    
        \includegraphics[scale=0.5]{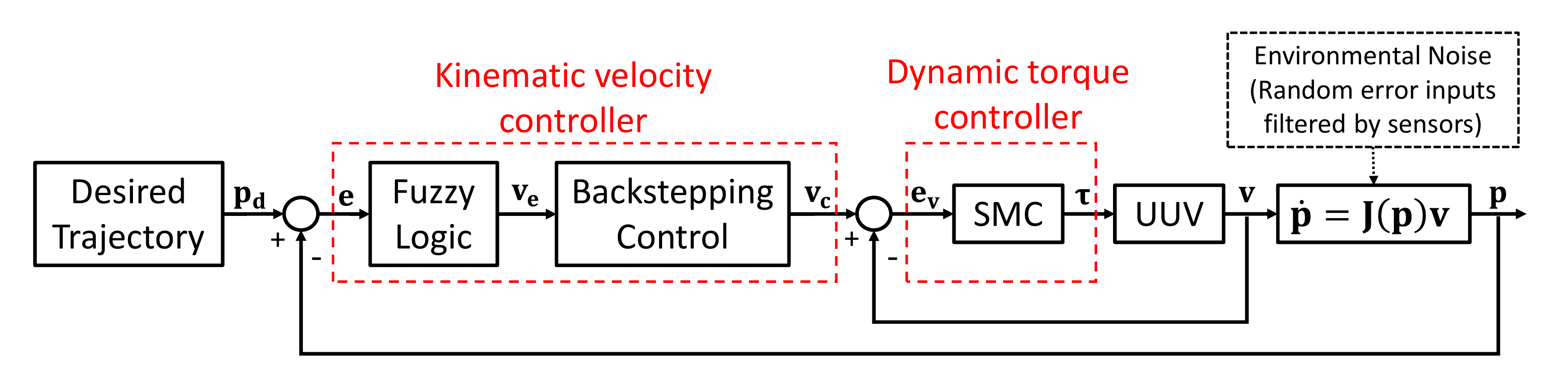}                         
        \caption*{Fig. 3. Schematic of the proposed trajectory tracking controller designed for the UUV}	
    
\end{center} 
\end{figure*}

\subsection{Fuzzy Logic System}
Suppose the maximum velocity vector of UUV is 
$\mathbf{v_m}=[v_{xm}~~v_{ym}~~v_{zm}~~v_{\psi m}]^T$, the input is the error between the desired and actual trajectories. For the fuzzification process, the input function is defined as
\begin{equation}
\mu_{i}=\frac{e(t)_{i}}{|e(t)_{i}|+1}\,,
\end{equation}
\noindent where $\mu_{i}$ is the input function of the $i^{th}$ axis, ranging between $x$, $y$, $z$, and $\psi$; $e(t)_{i}$ represents the error value at the time $t$ of the $i^{th}$ axis. The input function is chosen due to its smoother convergence tendency among the basic sigmoid functions ranging between -1 to 1, which can provide references of fewer fluctuations for the processed trajectory errors (see Fig. 4).

\begin{figure}[ht]
\begin{center}
    \includegraphics[scale=0.4]{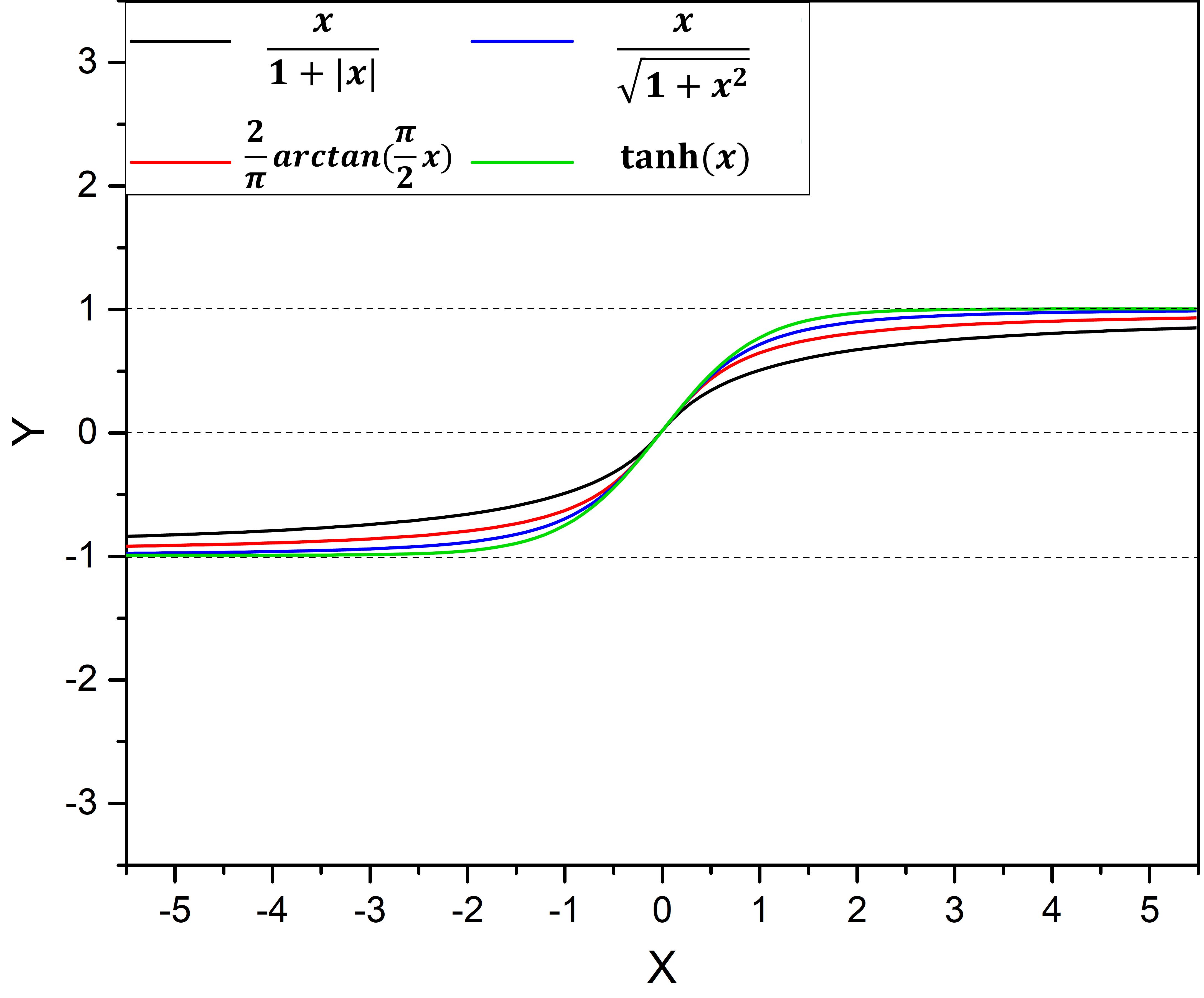}                         
    \caption*{Fig. 4. Some typical sigmoid functions ranging between -1 and 1.}	
\end{center} 
\end{figure}

Suppose the inference output is represented by a matrix $\mathbf{M_f}=[M_{fx}~~~M_{fy}~~~M_{fz}~~~M_{f\psi}]^T$. The fuzzy rules for the inference are defined as\\

\noindent IF $\lvert \mu_{i} \rvert\leq0.01$ , THEN the inference output $M_{fi}  =0$;\\
IF $0.01<\lvert \mu_{i} \rvert<0.99$, THEN the inference output $M_{fi}=\mu$;\\
IF $\lvert \mu_{i} \rvert \geq0.99$, THEN the inference output $M_{fi}  =sign(e_{i})$.\\

\noindent where $i$ referring to the $i^{th}$ axis of the UUV system.

Design a function on the fuzzy logic output $\mathbf{v_e}$ as
\begin{equation}
\mathbf{v_e}=\mathbf{M_f}\,\cdot\,\mathbf{v_m}\,,
\end{equation}
where $\mathbf{v_m}$ represents the maximum velocities of the UUV DOFs at their corresponding axes.

Therefore, when $\mathbf{e}(t)\rightarrow0$, $\mathbf{v_e}=\mathbf{M_f} \mathbf{v_m}\rightarrow0$; and when $\mathbf{e}(t)\rightarrow\infty$, $\mathbf{v_e}=\mathbf{M_f}\mathbf{v_m}\rightarrow\mathbf{v_m}$. The sigmoid function defined in Eq. (11) and the fuzzy rules restrict the inference output  within $[-1,1]$. Based on the restriction, the fuzzy logic limits its control outputs $\mathbf{v_e}\in[-\mathbf{v_m},\mathbf{v_m}]$.

The fuzzy logic output $\mathbf{v_e}$ shall not exceed the maximum possible value of the UUV velocity, and the definition of $\mu$ provides a smooth transition of the UUV at the beginning. Hence the speed-jump problem of the UUV can be alleviated. For the fuzzy logic outputs at the four degrees of freedom, the vector $\mathbf{v_e}$ can be written as
\begin{equation}
\mathbf{v_e}=
\begin{bmatrix}
   v_{ex}\\
   v_{ey}\\
   v_{ez}\\
   v_{e\psi}
\end{bmatrix}=
\begin{bmatrix}
   M_{fx}\,v_{xm}\\
   M_{fy}\,v_{ym}\\
   M_{fz}\,v_{zm}\\
   M_{f\psi}\,v_{\psi m}
\end{bmatrix}.
\end{equation}

\subsection{Fuzzy-refined Backstepping Component and Stability Analysis}
The two separately defined control laws are combined to obtain a better trajectory tracking control. The error variables in the control law of the backstepping method are replaced by the outputs of the fuzzy logic control in Eq. (13) \cite{r28}. The fuzzy logic outputs have alleviated the abrupt changes of the errors between the desired and actual trajectories. Therefore, the combined control law can be derived as follows
\begin{gather}
    \mathbf{v_c}=
\begin{bmatrix}
   u_c\\
   v_c\\
   w_c\\
   r_c
\end{bmatrix}\\ \notag
=\begin{bmatrix}
   k(v_{ex}\cos{\psi}+v_{ey}\sin{\psi})+u_d\cos{v_{e\psi}}-v_d\sin{v_{e\psi}}\\
    k(-v_{ex}\sin{\psi}+v_{ey}\cos{\psi})+u_d\sin{v_{e\psi}}-v_d\cos{v_{e\psi}}\\
    w_d+k_zv_{ez}\\
    r_d+k_\psi v_{e\psi} 
\end{bmatrix}.
\end{gather}
where $k$, $k_z$ and $k_\psi$ are positive constants.

Then the processed control velocities $\mathbf{v_c}$ are passed to the UUV, where they are calculated to keep pace with the desired trajectory through the dynamic model of the UUV.\\
\indent According to the Lyapunov stability theory, a special Lyapunov function $\Gamma_0$ is chosen as
\begin{equation}
\Gamma_0=\frac{1}{2}(e_x^2+e_y^2+e_z^2+e_\psi^2)\,.
\end{equation}

By Eq.s (1) and (14), the derivative of Eq. (15) can be obtained to prove the stability of the fuzzy logic-refined backstepping controller as \cite{r28}
\begin{equation}
    \begin{aligned}
\,\,\dot{\Gamma}_0=&e_x\,\dot{e}_x+e_y\,\dot{e}_y+e_z\,\dot{e}_z+e_\psi\,\dot{e}_\psi\\&
=e_x\,(\dot{x}_d-\dot{x})+e_y\,(\dot{y}_d-\dot{y})\\&\;\;\;
+e_z(\dot{z}_d-\dot{z})+e_\psi\,(\dot{\psi}_d-\dot{\psi})\\&
=e_x\,[(\cos{\psi_d}u_d-\sin{\psi_d}v_d)-(\cos\psi u_c-\sin{\psi} v_c)]\\&\;\;\;
+e_y\,[(\sin{\psi_d}u_d+\cos{\psi_d}v_d)-(\sin\psi u_c+\cos{\psi} v_c)]\\&\;\;\;
+e_z(w_d-w_c)+e_\psi\,(r_d-r_c)\\&
=e_x\,[(\cos{\psi_d}u_d-\sin{\psi_d}v_d)\\&\;\;\;
-(kv_{ex}+u_d(\cos{\psi}\cos{v_{e\psi}}-\sin{\psi}\sin{v_{e\psi}})\\&\;\;\;
+v_d(\sin{\psi}\cos{v_{e\psi}}-\cos{\psi}\sin{v_{e\psi}}))]\\&\;\;\;
+e_y\,[(\sin{\psi_d}u_d+\cos{\psi_d}v_d)\\&\;\;\;
-(kv_{ey}+u_d(\sin{\psi}\cos{v_{e\psi}}+\cos{\psi}\sin{v_{e\psi}})\\&\;\;\;
-v_d(\sin{\psi}\sin{v_{e\psi}}-\cos{\psi}\cos{v_{e\psi}}))]\\&\;\;\;
+e_z(-k_zv_{ez})+e_\psi\,(-k_\psi v_{e\psi})
\\&
\;\,\leq -ke_xv_{ex}-ke_yv_{ey}-k_ze_zv_{ez}-k_\psi e_\psi v_{e\psi}\,.
    \end{aligned}
\end{equation}

According to the definition of $\mathbf{v_{e}}$, $\mathbf{e}(t)$ and $\mathbf{v_{e}}$ are of the same sign. Because $\mathbf{v_{e}}$ is the same sign with $\mathbf{M_f}$ (Eq.s (12) and (13)); $\mathbf{M_f}$ is the same sign with $\mu_{i}$ according the inference rule and $\mu_{i}$ is the same sign with $\mathbf{e}(t)$ (Eq. (11)). Therefore, each multiplication of $\mathbf{e}(t)\mathbf{v_{e}}$ should be positive. Moreover, as $k$, $k_z$, $k_\psi$ are also positive constants, the result of Eq. (16) is believed to be less than and equal to zero, which demonstrates the stability of the designed fuzzy logic-based controller. 
\subsection{Sliding Mode Control Component and Stability Analysis}
To design the sliding mode control, the desired dynamics (s) should be introduced. Based on Eq. (2) where the vehicle system is of the second order for the velocity $\mathbf{v}$, the dynamics can be designed as
\begin{equation}
\mathbf{s}=\left[\frac{d}{dt}+\lambda\right]^2\,\int \mathbf{e_v} dt=\mathbf{\dot{e}_v}+2\lambda \mathbf{e_v}+\lambda^{2}\int \mathbf{e_v} dt,
\end{equation}
where $\frac{d}{dt}$ is the derivative operator; $\mathbf{e_v}$ represents the errors derived by the control velocities (see Fig. 3), $\mathbf{e_v}=\mathbf{v_c}-\mathbf{v}$; and $\lambda>0$ is a positive parameter \cite{r29}.\\
\indent Then take the derivative of $\mathbf{s}$, we can get
\begin{equation}
\mathbf{\dot{s}}=\mathbf{\ddot{e}_v}+2\lambda \mathbf{\dot{e}_v}+\lambda^{2} \mathbf{e_v}\,,
\end{equation}
where $\mathbf{\dot{e}_v}=\mathbf{\dot{v}_c}-\mathbf{\dot{v}}$ .

To keep the system states behave consistently with the desired dynamics, the derivative in Eq. (13) should be equal to zero. This means the system states are on the sliding surface of the perfect tracking. At the same time, plug in the equation of the UUV dynamic model (Eq. (2)),
\begin{eqnarray}
&&\;\;\;\mathbf{\dot{s}}=\mathbf{\ddot{e}_v}+2\lambda \mathbf{\dot{e}_v}+\lambda^{2} \mathbf{e_v}=0 \notag\\
&& \mathbf{\ddot{e}_v}+2\lambda (\mathbf{\dot{v}_c}-\mathbf{\dot{v}})+\lambda^{2} \mathbf{e_v}=0 \notag\\
&& \mathbf{\ddot{e}_v}+2\lambda (\mathbf{\dot{v}_c}-(\boldsymbol{\uptau}-\mathbf{C}\mathbf{v}-\mathbf{D}\mathbf{v}-\mathbf{g})\mathbf{M}^{-1})+\lambda^{2} \mathbf{e_v}=0 \notag\\
&& \boldsymbol{\uptau}=\mathbf{M}(\mathbf{\dot{v}_c}+\frac{\mathbf{\ddot{e}_v}}{2\lambda}+\frac{\lambda}{2}\mathbf{e_v})+\mathbf{C}\mathbf{v}+\mathbf{D}\mathbf{v}+\mathbf{g}\,.
\end{eqnarray}

The standard sliding mode control law is defined as
\begin{equation}
    \boldsymbol{\uptau}=\boldsymbol{\hat{\uptau}}+\boldsymbol{\uptau_c}\,,
\end{equation}
where $\boldsymbol{\hat{\uptau}}$ represents the major control law, which is continuous and model-based. It is designed to maintain the trajectory consistently on the sliding surface. $\boldsymbol{\uptau_c}$ represents the switching control law, dealing with the model uncertainty. When the trajectory is getting out of control, $\boldsymbol{\uptau_c}$ is used to push the trajectory back to the sliding surface and keep the good tracking. For Eq. (14), supposing a simplification $\mathbf{\ddot{e}_v} \approx -k \mathbf{\dot{e}_v}$ based on the error acceleration feedback control to reduce computation complexity, the estimated item in the major control law $\boldsymbol{\hat{\uptau}}$ can be deducted as 
\begin{equation}
\boldsymbol{\hat{\uptau}}=\mathbf{\hat{M}}(\mathbf{\dot{v}_c}+\frac{-k \mathbf{\dot{e}_v}}{2\lambda}+\frac{\lambda}{2}\mathbf{e_v})+\mathbf{\hat{C}}\mathbf{v}+\mathbf{\hat{D}}\mathbf{v}+\mathbf{\hat{g}}\,,
\end{equation}
where $\hat{\mathbf{M}}$,  $\hat{\mathbf{C}}$,  $\hat{\mathbf{D}}$,  $\hat{\mathbf{g}}$ are the estimated values of $\mathbf{M}$, $\mathbf{C}$, $\mathbf{D}$ and $\mathbf{g}$, where approximate values can be obtained from the practical case respectively \cite{r25}.

The switching item $\boldsymbol{\uptau_c}$ in sliding mode control can be defined as
\begin{equation}
\boldsymbol{\uptau_c}=-\mathbf{K}_1\mathbf{s}-\mathbf{K}_2|s|^{r}sign(\mathbf{s})\,,
\end{equation}
where $sign(\mathbf{s})$ is the nonlinear sign function of $\mathbf{s}$, $\mathbf{K}_1$ and $\mathbf{K}_2$ are positive coefficients. $\mathbf{K}_1\geq \eta +F$ and $\mathbf{K}_2\geq \eta +F$,
where $\eta$ is the design parameter which is always chosen as a positive constant. $F$ represents the upper bound of the difference between the system actual output $f(\mathbf{v})$ and the system output estimation $\hat{f}(\mathbf{v})$ in the following form as
\begin{equation}
F=|f(\mathbf{v})-\hat{f}(\mathbf{v})|.
\end{equation}

Additionally, an adaptive variation term $\widetilde{\mathbf{\uptau}}_{est}$ is added to the control law, where $\dot{\widetilde{\mathbf{\uptau}}}_{est}=\Gamma\mathbf{s}$ and $\Gamma$ represents a positive constant. Hence the final sliding mode control law is defined as
\begin{gather}
    \boldsymbol{\uptau}=\boldsymbol{\hat{\uptau}}+\widetilde{\mathbf{\uptau}}_{est}+\mathbf{\uptau_c} \notag\\
    =\mathbf{\hat{M}}(\mathbf{\dot{v}_c}+\frac{-k \mathbf{\dot{e}_v}}{2\lambda}+\frac{\lambda}{2}\mathbf{e_v})+\mathbf{\hat{C}}\mathbf{v}+\mathbf{\hat{D}}\mathbf{v}+\mathbf{\hat{g}} \notag \\ +\widetilde{\mathbf{\uptau}}_{est}-\mathbf{K}_1\mathbf{s}-\mathbf{K}_2|s|^{r}sign(\mathbf{s})\,.
\end{gather}

To prove the stability of the SMC, construct a Lyapunov function as
\begin{equation}
    \mathbf{V}=\frac{1}{4\lambda}\mathbf{s}^{T}\mathbf{M}\mathbf{s}+\frac{1}{2}\mathbf{Q}^{T}\Gamma^{-1}\mathbf{Q}\,,
\end{equation}
where $\mathbf{Q}=\widetilde{\mathbf{\uptau}}_r-\widetilde{\mathbf{\uptau}}_{est}$ and $\widetilde{\mathbf{\uptau}}_r=\mathbf{\widetilde{M}}\mathbf{\dot{v}}_r+\mathbf{\widetilde{C}}\mathbf{v}_r+\mathbf{\widetilde{D}}\mathbf{v}+\mathbf{\widetilde{g}}$.

Previously we have given $\mathbf{e_v}=\mathbf{v_c}-\mathbf{v}$, and $\mathbf{s}
\mathbf{\dot{e}_v}+2\lambda \mathbf{e_v}+\lambda^{2}\int \mathbf{e_v} dt$, such that two equations can be deducted as
\begin{equation}
    \mathbf{v}=\mathbf{v_c}-\frac{\mathbf{s}-\mathbf{\dot{e}_v}-\lambda^{2}\int \mathbf{e_v} dt}{2\lambda},
\end{equation}
\begin{equation}
    \mathbf{\dot{v}}=\mathbf{\dot{v}_c}-\frac{\mathbf{\dot{s}}-\mathbf{\ddot{e}_v}-\lambda^{2}\mathbf{e_v}}{2\lambda},
\end{equation}
therefore the following items can be defined as
\begin{equation}
    \mathbf{v_r}=\mathbf{v_c}+\frac{\mathbf{\dot{e}_v}+\lambda^{2}\int \mathbf{e_v} dt}{2\lambda},
\end{equation}
\begin{equation}
    \mathbf{\dot{v}_r}=\mathbf{\dot{v}_c}+\frac{\mathbf{\ddot{e}_v}+\lambda^{2}\mathbf{e_v}}{2\lambda}.
\end{equation}

By substituting into Eq. (2),
\begin{gather}
    \mathbf{M}\frac{\mathbf{\dot{s}}}{2\lambda}+\mathbf{C}\frac{\mathbf{s}}{2\lambda} \notag \\
    =\mathbf{M}(\mathbf{\dot{v}_c}+\frac{\mathbf{\ddot{e}_v}+\lambda^{2}\mathbf{e_v}}{2\lambda})+\mathbf{C}(\mathbf{v_c}+\frac{\mathbf{\dot{e}_v}+\lambda^{2}\int \mathbf{e_v} dt}{2\lambda}) \notag\\
    +\mathbf{D}\mathbf{v}+g-\mathbf{\uptau}
    =\mathbf{M}\mathbf{\dot{v}_r}+\mathbf{C}\mathbf{v_r}+\mathbf{D}\mathbf{v}+g-\mathbf{\uptau}.
\end{gather}

Based on previous definitions, the derivative of Eq. (27) can be simplified as
\begin{eqnarray}
&&\mathbf{\dot{V}}=\frac{1}{4\lambda}(\mathbf{s}^{T}\mathbf{\dot{M}}\mathbf{s}+\mathbf{\dot{s}}^{T}\mathbf{M}\mathbf{s}+\mathbf{s}^{T}\mathbf{M}\mathbf{\dot{s}}) \notag \\
&&+\frac{1}{2}\mathbf{\dot{Q}}^{T}\Gamma^{-1}\mathbf{Q}+\frac{1}{2}\mathbf{Q}^{T}\Gamma^{-1}\mathbf{\dot{Q}} \notag\\
&& =\frac{1}{2\lambda}\mathbf{s}^{T}(\mathbf{M}\mathbf{\dot{s}}+\mathbf{C}\mathbf{s})+\frac{1}{2}\mathbf{\dot{Q}}^{T}\Gamma^{-1}\mathbf{Q}+\frac{1}{2}\mathbf{Q}^{T}\Gamma^{-1}\mathbf{\dot{Q}} \notag\\
&& = \mathbf{s}^{T}(\mathbf{M}\mathbf{\dot{v}_r}+\mathbf{C}\mathbf{v_r}+\mathbf{D}\mathbf{v}+g-\mathbf{\uptau})+\mathbf{\dot{Q}}^{T}\Gamma^{-1}\mathbf{Q}. 
\end{eqnarray}

By substituting Eq. (19),
\begin{gather}
\mathbf{\dot{V}}
= \mathbf{s}^{T}(\mathbf{M}\mathbf{\dot{v}_r}+\mathbf{C}\mathbf{v_r}+\mathbf{D}\mathbf{v}+g-\mathbf{\uptau}) \notag\\
+(\dot{\widetilde{\mathbf{\uptau}}}_r-\dot{\widetilde{\mathbf{\uptau}}}_{est})^{T}\Gamma^{-1}\mathbf{Q} \notag\\
=-\mathbf{s}^{T}(\mathbf{K}_1\mathbf{s}+\mathbf{K}_2|\mathbf{s}|^{r}sign(\mathbf{s}))+(\dot{\widetilde{\mathbf{\uptau}}}_r)^{T}\Gamma^{-1}\mathbf{Q}.  
\end{gather}

The dynamic item $\mathbf{\widetilde{\uptau}}_r$ is bounded due to the slow velocity of the underwater vehicle and $\mathbf{s}^{T}(\mathbf{K}_1\mathbf{s}+\mathbf{K}_2|\mathbf{s}|^{r}\rm{sign}(\mathbf{s}))\geq(\dot{\widetilde{\mathbf{\uptau}}}_r)^{T}\Gamma^{-1}\mathbf{Q}$. When $\mathbf{K}_1$, $\mathbf{K}_2$ and $\Gamma$ are assigned with large enough values at the design step, $\mathbf{\dot{V}}\leq 0$ can be achieved and $\mathbf{V}$ is ensured to be bounded, thus leading to the conclusion that $\mathbf{Q}$ is bounded. Then design a new Lyapunov function as
\begin{equation}
    \mathbf{V}_2=\frac{1}{4\lambda}\mathbf{s}^{T}\mathbf{M}\mathbf{s},
\end{equation}
whose derivative can be deducted as
\begin{equation}
    \mathbf{\dot{V}}_2=\mathbf{s}^{T}(\mathbf{Q}-\mathbf{K}_1\mathbf{s}-\mathbf{K}_2|s|^{r}sign(s)),
\end{equation}
where  $0 < r < 1$. Suppose $||Q|| < a$,
\begin{equation}
    \mathbf{\dot{V}}_2\leq \frac{1}{2}||s||^2+\frac{1}{2}a-\lambda_{min}(\mathbf{K}_1)||s||^2-\lambda_{min}(\mathbf{K}_2)||s||^{1+r},
\end{equation}
choose $\mathbf{K}_1$ when $\lambda_{min}(\mathbf{K}_1) > \frac{1}{2}+\beta$, where $\beta > 0$,
\begin{equation}
    \mathbf{\dot{V}}_2\leq -\beta||s||^2-\lambda_{min}(\mathbf{K}_2)||s||^{1+r}+\frac{1}{2}a,
\end{equation}
which induces that the Lyapunov function converges to a range close to zero in a finite time and $\mathbf{s}$ converges to a range close to zero in a finite time. Therefore, the supposed condition of the Lyapunov theorem can be regarded as satisfied, thus proving the stability of the designed SMC. 

\section{SIMULATION RESULTS AND ANALYSIS }
In this section, simulation results of the FBSTT are presented and analyzed. The desired trajectory is given in the form of 3D helix movement. The tracking trajectories, errors and normalized forces of the conventional backstepping combined with the SMC trajectory tracking (BSTT) method and the FBSTT method are compared in the simulation. The simulation involving disturbance that imitates the environmental noise has also been illustrated. Moreover, the running time of the conventional method (BSTT), the FBSTT and the MPC are compared to show the efficiency respectively.
%
%
\subsection{3D Helix Tracking}
In this section, the 3D helix serves as the desired trajectory for the FBSTT and BSTT methods and meanwhile the physical constraints of the UUV torques are applied to both tracking methods. The required torques output by the dynamic torque controller $\boldsymbol{\uptau}$ are restricted by the applied practical constraints  $\boldsymbol{\uptau_c}$. This affects the tracking effectiveness as the vehicle possibly cannot provide the excessive torques required by the controller. The FBSTT method focuses on restricting the dynamic outputs (torques/forces) given by the controller within the range of the constraints, which is more applicable in practical cases of the UUV. 

The initial position of the desired trajectory is set to be $(0,0,0,0)$, while the initial position of the control trajectory is set to be $(0,-10,0,0)$. The difference between the initial positions is given to test the correcting ability of the two tracking strategies when certain amount of deviation is applied at the beginning of an axis, i.e. the y axis. Assuming the desired trajectory of the UUV is $x_d=10 \sin{0.1t}$, $y_d=10-10\cos{0.1t}$, $z_d=0.5t$ and $\psi_d=0.1t$.

The range of the controller parameters are chosen based on the references of the UUV parameters in actual cases \cite{r8}. The choice of $K$ had better not be too large to introduce the chattering problem of the SMC and not too small to drag the speed of reaching the sliding mode surface \cite{r30,r31}. Specific values of the parameter constants are determined based on trial and error, where $K_1$ and $K_2$ are chosen as 50. $k$ is 2.5, $k_z$ and $k_\psi$ are assigned as 1, as the small design parameters slow the tracking speed while the large design parameters cannot promise the smooth approaching tendency of the FBSTT method at the initial state (see Fig. 5).
\begin{figure}[ht]
\begin{center}
        \includegraphics[width=0.49\textwidth]{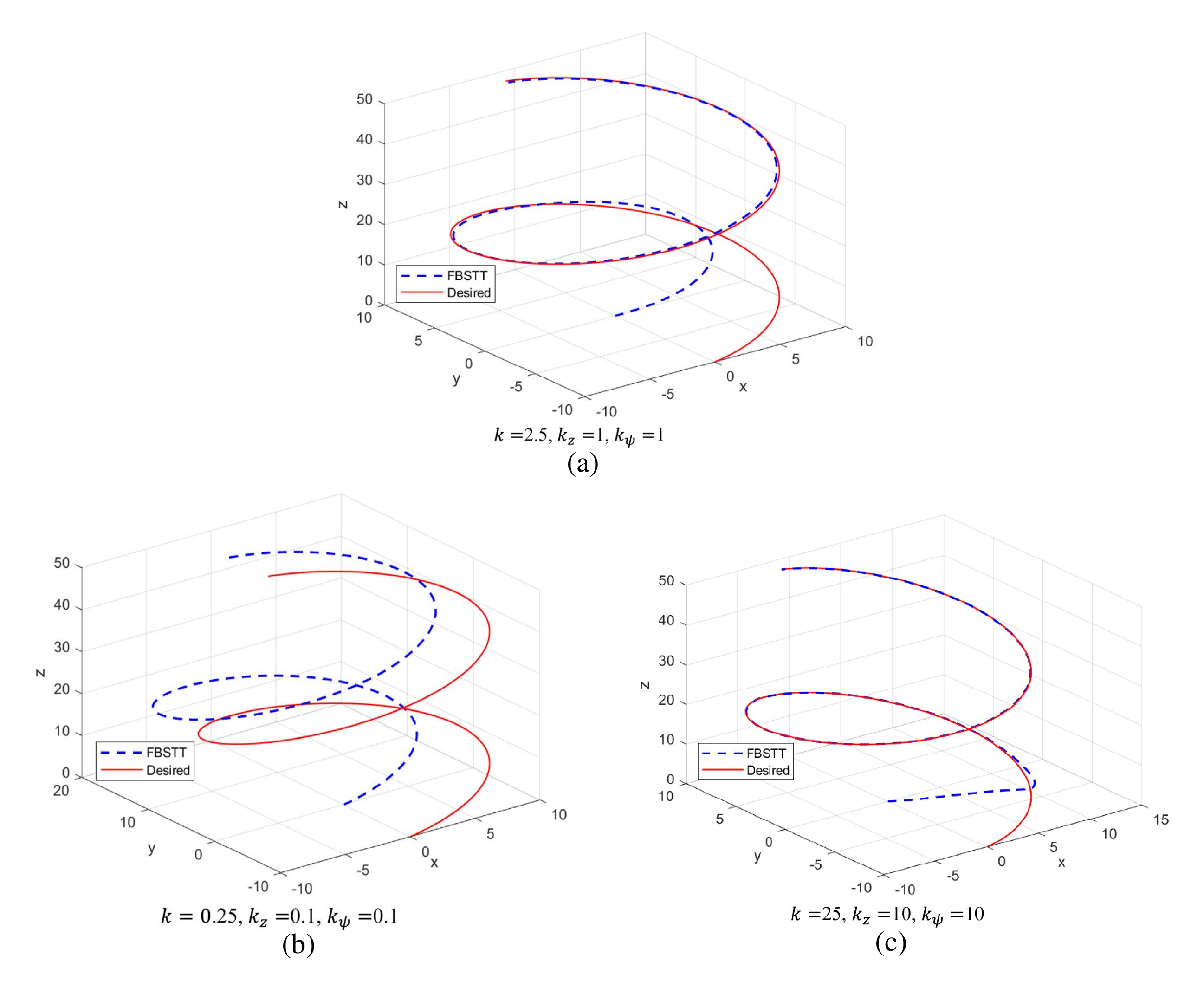}                         
        \caption*{Fig. 5. Comparison of 3D helix tracking using the FBSTT control under different design parameters}	
\end{center} 
\end{figure}

\begin{figure}[ht]
\begin{center}
        \includegraphics[width=0.49\textwidth]{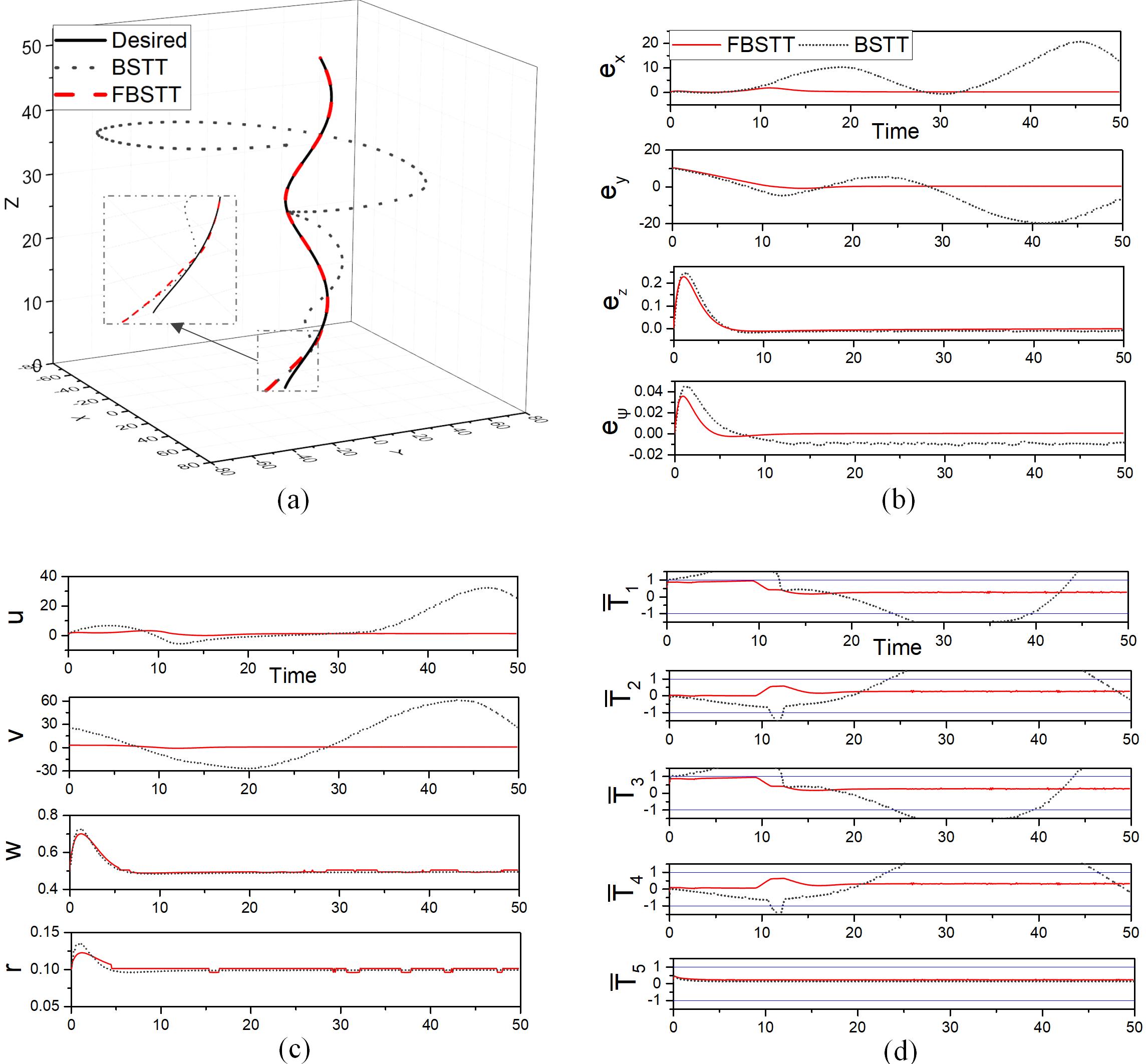}                         
        \caption*{Fig. 6. 3D helix tracking using the FBSTT control and the BSTT control considering torques’ constraints. (a) Comparison of tracking trajectories, (b) Comparison of trajectory errors, (c) Comparison of control velocities, (d) Comparison of normalized forces.}	
\end{center} 
\end{figure}

The FBSTT control (in red) has eliminated the error and achieved the same trajectory with the desired helix at the end (see Fig. 6(a)). An smooth approaching curve to the desired trajectory at the beginning is presented. On comparison, the trajectory controlled by the BSTT method (in black dots) fails to catch up with the desired helix at the beginning and throughout the whole process. This is due to the jumping speeds required by the backstepping method, which turns into excessive torques when passing through the dynamic model. In the BSTT results, the constraints in this simulation limits the vehicle to offer enough torques to push itself back to the desired trajectory when the deviation occurs. Therefore in BSTT simulation, the deviation accumulates increasingly and performs a messy curve at the end. Meanwhile, the FBSTT control successfully restricts its required dynamic ouputs within the constraints, which results into a satisfactory tracking performance. This conclusion is also supported by the error and control velocity curves given in Fig.s 6(b) and (c). 

\begin{table}[ht]
\centering
\captionsetup{justification=centering}
\caption*{TABLE \uppercase\expandafter{\romannumeral1}. Maximum Velocities of the FBSTT Control and the BSTT Control Considering the Dynamic Constraints}
\begin{tabular}{|c|c|c|c|c|}
\hline
& $u_c$ (m/s) & $v_c$ (m/s) & $w_c$ (m/s) & $r_c$ (m/s)\\
\hline
FBSTT & 2.9535 & 2.2922 & 0.6965 & 0.1214\\
\hline
BSTT & 220.8912 & -319.6140 & 0.7248 & 0.1347\\
\hline
Desired & 1 & 0 & 0.5 & 0.1\\
\hline
\end{tabular}
\end{table}

The fluctuation of errors of the FBSTT control (in red) and the BSTT control (in black dots) are compared in Fig. 6(b). For the FBSTT control, in the x, z and $\psi$ axes, the errors are all eliminated to approximate zero at the end. In the y axis who has the largest initial deviation, the error curve of the FBSTT control shows a smooth tendency when converging, and it finally sustains at zero. At the same time, for the BSTT method, large fluctuations are presented, and the convergence of errors is not performed throughout the whole process, i.e. in x and y axes. The performance of the BSTT method proves that it cannot eliminate the velocity errors in most of the states when there is torques’ constraints applied, thus leading to the failure of tracking the desired trajectory shown in Fig. 6(a). Similar conclusion can be derived by the control velocities presented in Fig. 6(c), the BSTT method performs larger fluctuations in x and y axes, and never coincides with the desired velocity throughout the whole process. Its control velocity at y axis reaches the maximum of -319.614m/s, which is dramatically away from the desired velocity of 0m/s (see TABLE \uppercase\expandafter{\romannumeral1}). On comparison, the FBSTT control offers a more reliable tracking result, with smaller velocity fluctuations and better capability of eliminating the errors during the tracking process.

\begin{table}[ht]
\centering
\captionsetup{justification=centering}
\caption*{TABLE \uppercase\expandafter{\romannumeral2}. Maximum Normalized Forces Required at the Five thrusters Using the FBSTT Control and the BSTT Control Considering the Dynamic Constraints}
\begin{tabular}{|c|c|c|c|c|c|}
\hline
& $\overline{T}_1$ & $\overline{T}_2$ & $\overline{T}_3$ & $\overline{T}_4$ & $\overline{T}_5$\\
\hline
FBSTT & 0.9323 & 0.5385 & 0.8980 & 0.5727 & 0.3676\\
\hline
BSTT & -52.8408 & -30.5963 & -52.8754 & -30.5626 & 0.3797\\
\hline
\end{tabular}
\end{table}

Required forces of the five thrusters are normalized in Fig. 6(d) to show the practical performance. The results of the comparison also prove that the BSTT method (in black dots) largely exceeds the constraints of some thrusters and larger fluctuations of the forces are created accordingly. Especially for thrusters $T_1$ and $T_2$, abrupt changes are produced at the initial status; large fluctuations that exceeds the constraints exist throughout the whole process. For example, the maximum value of normalized forces required by $\overline{T}_1$ and $\overline{T}_2$ are -52.8408 and -30.5963, which are impossible to achieve within the constraints of the vehicle (see TABLE \uppercase\expandafter{\romannumeral2}). At the same time, the FBSTT control (in dotted black) successfully limits the required forces in a dramatic small range and meanwhile within the range of the constraints, such that smooth force curves for all the thrusters are performed.

\subsection{3D Helix Tracking Involving Noise} 
In this section, the environmental disturbance is considered in the 3D helix tracking simulation the FBSTT and the BSTT methods, whose results are shown and compared in the following figures. The environmental disturbance is addressed by introducing a random error input at the status of forming the trajectory as $rand[-0.1,0.1]$, and then filtered by the virtual sensors to imitate the UUV tracking condition in practical applications.

\begin{figure}[ht]
\begin{center}
    
        \includegraphics[width=0.49\textwidth]{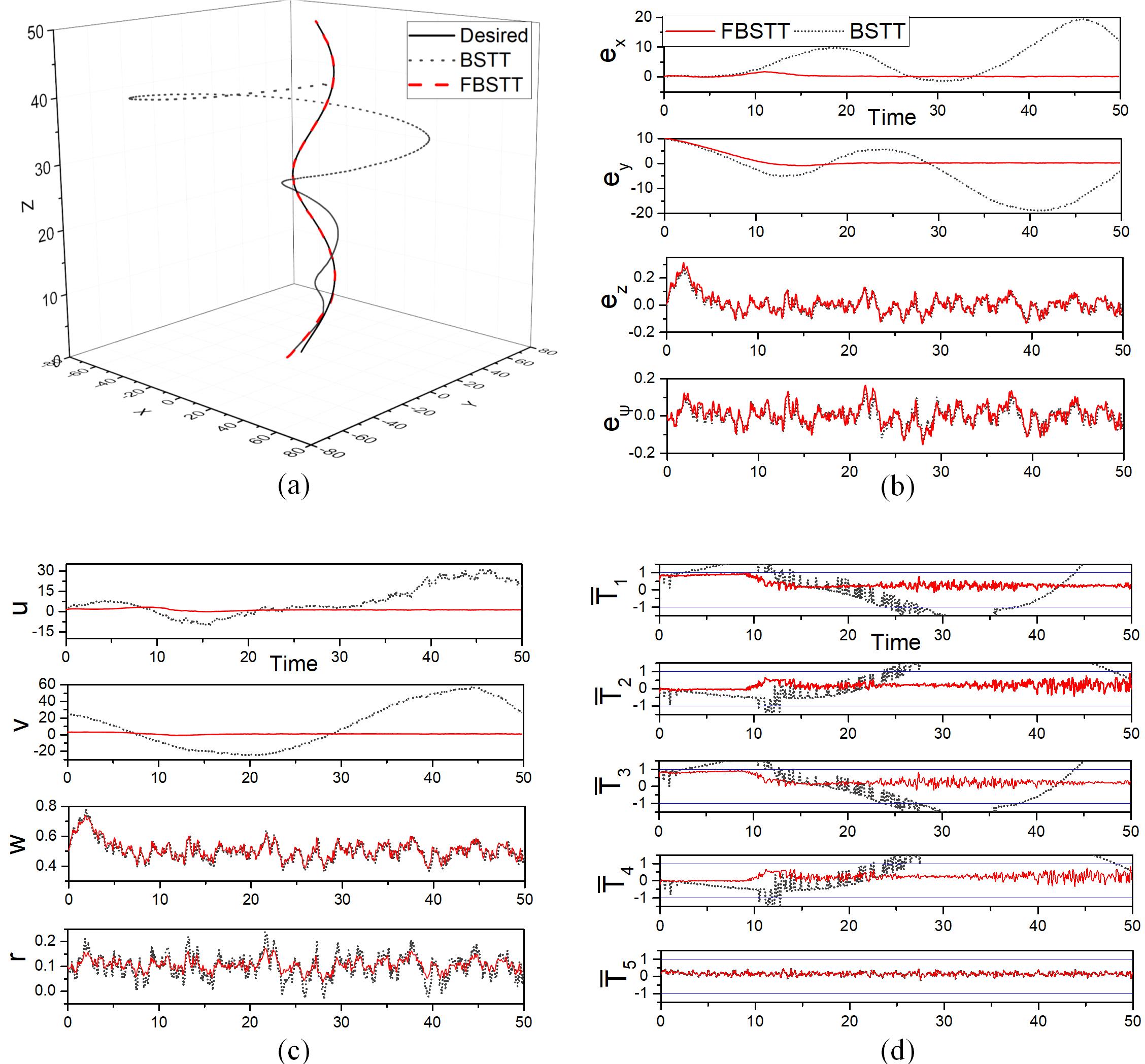}                         
        \caption*{Fig. 7. 3D helix tracking using the FBSTT control and the BSTT control with noise involved. (a) Comparison of tracking trajectories, (b) Comparison of trajectory errors, (c) Comparison of control velocities, (d) Comparison of normalized forces.}
    
\end{center} 
\end{figure}

Both trajectories of the FBSTT control (in red) and the BSTT control (in black dots) perform exactly the same tendency with the trajectory shown in Fig. 6(a), where the BSTT trajectory presents an abrupt turning at the initial state and produces an increasing deviation throughout the whole process, while the FBSTT trajectory achieves a satisfactory tracking result with a smooth tendency when approaching the desired trajectory. This demonstrates the effect of environmental disturbance does not produce deviations for the FBSTT tracking trajectory, which supports the robustness of the proposed control. The conclusion can also be derived based on the error and control velocity curves given in Fig.s 7(b) and (c).

Compared to the FBSTT control, error curves of the BSTT control (in black dots) change more abruptly especially in x and y axis, at the same time, the errors of FBSTT control (in red) are eliminated in a faster and smoother manner though small chattering brought by the simulated disturbance are presented (see Fig. 7(b)). Similarly, for the control velocities of both methods shown in Fig. 7(c), velocities produced by the conventional backstepping method never approach the desired velocity at x and y axis with more violent fluctuations while the FBSTT control sustains at the desired velocity throughout the tracking procedure. The sharper jumps of the backstepping method create higher demands for the control velocities at the beginning, therefore, the forces required by the vehicle for producing the corresponding velocities exceed the constraints of the vehicle (Fig. 7(d) and TABLE \uppercase\expandafter{\romannumeral3}).
\begin{table}[ht]
\centering
\captionsetup{justification=centering}
\caption*{TABLE \uppercase\expandafter{\romannumeral3}. Maximum Normalized Forces Required at the Five thrusters Using the FBSTT Control and the BSTT Control Involving the Noise}
\begin{tabular}{|c|c|c|c|c|c|}
\hline
& $\overline{T}_1$ & $\overline{T}_2$ & $\overline{T}_3$ & $\overline{T}_4$ & $\overline{T}_5$\\
\hline
FBSTT & 0.9655 & 0.8820 & 0.9088 & 0.8377 & 0.4486\\
\hline
BSTT & -38.5109 & -25.3069 & -38.4884 & -25.2449 & 0.4660\\
\hline
\end{tabular}
\end{table}

\subsection{Efficiency Evaluation}
Compared with popular methods applied in the field of trajectory tracking, such as the MPC, the FBSTT method performs higher efficiency with less running time required (see Fig. 8). Thanks to the low complexity of the fuzzy logic defined in this paper, it successfully limits the extra burden added on the computation process of BSTT in an extremely small range. Therefore the FBSTT method sustains almost the same running time as the BSTT method even after four times of iteration. While the computation of the MPC always consumes the longest time. The increment of the iteration does not affect the efficiency superiority of the FBSTT, thus proving its effectiveness especially in the UUV trajectory tracking, who prefers instant feedback given by the algorithm. 

The FBSTT method has processed the velocity errors with a fuzzy logic, thus passing error curves with smaller fluctuations to the dynamic model of the UUV. Therefore the SMC applied in the dynamic model computes the input data and all the other impactors from environment with less abruptness and higher accuracy. Based on the design, the SMC derives smaller amounts and smoother changes for the torques, and a more reliable reference for eliminating the tracking errors is obtained. The smoothness and robustness of the FBSTT control promise the trajectory tracking results of the UUV can be maintained in a satisfactory mode, even in a complex underwater environment.
\begin{figure}[ht]
\begin{center}
    
        \includegraphics[scale=0.65]{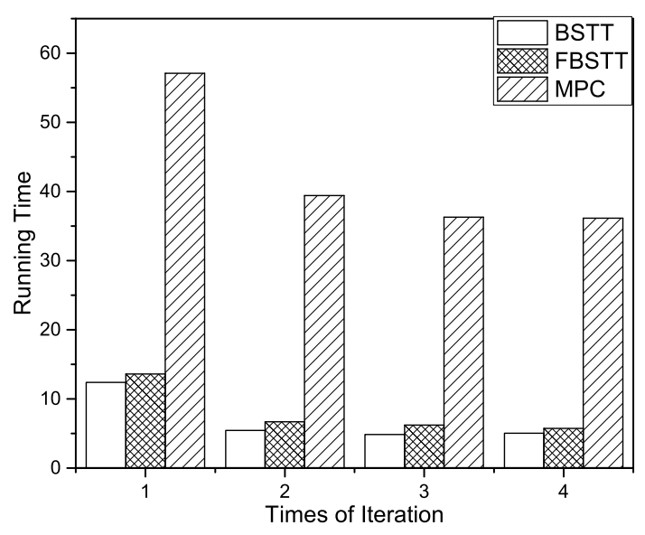}                         
        \caption*{Fig. 8. Running time comparison of the BSTT, FBSTT and MPC methods for 3D helix tracking of 100 seconds.}
    
\end{center} 
\end{figure}

\section{Conclusion}
In this paper, the actuator saturation problem in the UUV trajectory tracking control are resolved by a fuzzy logic-based cascade control strategy. The actuator saturation problem appears at the dynamic component of the UUV, induced by the speed jumps exist at the kinematic component, which usually occurs in conventional control methods such as the backstepping method. The cascade control strategy uses the fuzzy logic to alleviate the excessive speed references, and restricts the dynamic outputs (forces) in acceptable domains. Instead of directly based on the errors between desired trajectory and actual trajectory, the fuzzy logic introduced in the FBSTT method processes errors with a sigmoid function to reduce fluctuations, and the decision algorithm in the fuzzy logic helps to limit the errors in a certain range. The processed errors are passed to the SMC component to achieve rational torques/forces to obtain satisfactory trajectory tracking results. The 3D helix simulation results of the comparison between the FBSTT control and the conventional backstepping method-based control verifies the efficiency and accuracy of the FBSTT control when dynamic constraints are applied. Moreover, the environmental noise is considered in the simulation section to further verify the effectiveness of the proposed control in actual cases. In the future study, the FBSTT strategy is supposed to be applied in practical experiments to test whether it corresponds the results shown in the simulation, with more complex environmental factors such as the effect of ocean currents involved.

\ifCLASSOPTIONcaptionsoff
  \newpage
\fi



%

\scriptsize
\bibliographystyle{IEEEtran}
\bibliography{ref}



%









\end{document}